\documentclass[10pt,twocolumn]{article}

\usepackage{xcolor}
\usepackage{multirow}
\usepackage{graphicx}
\usepackage{amsmath}
\usepackage{amssymb}
\usepackage{fontspec}
\usepackage{amsmath, amssymb, amsthm}  
\usepackage{graphicx}                  
\usepackage{subfig}                    
\usepackage{booktabs}                  
\usepackage{algorithm}                 
\usepackage{algorithmic}               
\usepackage{multirow}                  
\usepackage{xcolor}                    
\usepackage{hyperref}                  
\usepackage{cite}                      
\usepackage{balance}                   
\usepackage{flushend}                  
\usepackage{geometry}                  
\usepackage{caption}                   
\usepackage{float}                     
\usepackage{titlesec}                  
\usepackage{abstract}                  
\usepackage{lipsum}                    

\titlespacing*{\section}{0pt}{12pt}{6pt}
\titlespacing*{\subsection}{0pt}{10pt}{4pt}
\titlespacing*{\subsubsection}{0pt}{8pt}{2pt}

\newcommand{\vect}[1]{\boldsymbol{#1}}  
\newcommand{\reals}{\mathbb{R}}         

\begin{document}

\title{\Large \textbf{GEnSHIN: Graphical Enhanced Spatio-temporal Hierarchical Inference Network for Traffic Flow Prediction}}

\author{
    \textbf{Zhiyan Zhou}$^{1}$\thanks{Corresponding author: 202311079187@mail.bnu.edu.cn},
    \textbf{Junjie Liao}$^{1}$\thanks{These authors contributed equally to this work.},
    \textbf{Manho Zhang}$^{1}$,
    \textbf{Yingyi Liao}$^{1}$,
    \textbf{Ziai Wang}$^{1}$\\
    $^{1}$Beijing Normal University
}

\maketitle

\begin{abstract}
\noindent 
\textbf{Abstract} --
With the acceleration of urbanization, intelligent transportation systems have an increasing demand for accurate traffic flow prediction. This paper proposes a novel Graph Enhanced Spatio-temporal Hierarchical Inference Network (GEnSHIN) to handle the complex spatio-temporal dependencies in traffic flow prediction. The model integrates three innovative designs: 1) An attention-enhanced Graph Convolutional Recurrent Unit (GCRU), which strengthens the modeling capability for long-term temporal dependencies by introducing Transformer modules; 2) An asymmetric dual-embedding graph generation mechanism, which leverages the real road network and data-driven latent asymmetric topology to generate graph structures that better fit the characteristics of actual traffic flow; 3) A dynamic memory bank module, which utilizes learnable traffic pattern prototypes to provide personalized traffic pattern representations for each sensor node, and introduces a lightweight graph updater during the decoding phase to adapt to dynamic changes in road network states. Extensive experiments on the public dataset METR-LA show that GEnSHIN achieves or surpasses the performance of comparative models across multiple metrics such as Mean Absolute Error (MAE), Root Mean Square Error (RMSE), and Mean Absolute Percentage Error (MAPE). Notably, the model demonstrates excellent prediction stability during peak morning and evening traffic hours. Ablation experiments further validate the effectiveness of each core module and its contribution to the final performance.

\vspace{0.1cm}
\end{abstract}

\section{Introduction}
\label{sec:introduction}

As a core task of Intelligent Transportation Systems (ITS), traffic flow prediction is of paramount importance for alleviating urban congestion, optimizing route planning, and controlling environmental pollution \cite{li2018diffusion,ha2011traffic,kipf2016semi}. However, accurate traffic prediction is highly challenging \cite{ha2011traffic}, primarily due to the complex spatio-temporal dependencies of traffic data: in the temporal dimension, flow exhibits highly nonlinear and dynamic characteristics (e.g., morning/evening peaks, periodic weekend effects); in the spatial dimension, the road network topology is intricate, with explicit physical connections and implicit mutual influences between upstream and downstream road segments.

Traditional time series analysis methods (e.g., ARIMA, VAR) typically rely on stationarity assumptions and struggle to effectively capture nonlinear dynamic features in the data \cite{guo2019attention}. With the advancement of deep learning, Convolutional Neural Networks (CNN) and Recurrent Neural Networks (RNN) have been widely applied to traffic prediction. However, standard CNNs have difficulty directly processing traffic road network data with Non-Euclidean structures, while RNNs often face gradient vanishing or explosion issues when dealing with long sequences. Furthermore, these methods often treat temporal and spatial features separately, failing to fully achieve joint modeling of spatio-temporal dimensions.

In recent years, Graph Neural Networks (GNNs) have shown great potential in traffic prediction due to their natural advantage in handling graph-structured data \cite{wu2019graph}. Building on this, researchers have derived various prediction models based on Spatio-Temporal Graph Neural Networks (ST-GNNs), effectively improving prediction accuracy by combining graph convolution with sequence models.

Despite significant progress by existing Graph Convolutional Network (GCN)-based methods, two main limitations remain: (1) \textbf{Dependency on prior graph construction}: Most rely on predefined static graph structures (e.g., adjacency matrices based on distance or connectivity), ignoring data-driven latent dynamic topological relationships; (2) \textbf{Neglect of node heterogeneity}: They usually employ parameter-shared graph convolution kernels, making it difficult to capture the unique traffic patterns of sensor nodes at different geographic locations. To address these issues, this paper proposes a Graphical Enhanced Spatiotemporal Hierarchical Inference Network (GEnSHIN). The main contributions of this paper are summarized as follows:

\begin{itemize}
    \item \textbf{Attention-enhanced GCRU unit}: Drawing on the design ideas of DGCRN \cite{li2022dynamic}, graph convolution operations replace the linear transformations in traditional RNNs to achieve deep integration of spatial topological information and temporal evolution; Transformer modules are introduced to further enhance the model's ability to capture long-term dependencies.

    \item \textbf{Asymmetric dual-embedding graph generation mechanism}: Two independent learnable node embedding vectors are designed to generate an adaptive graph structure via asymmetric matrix multiplication, which is then weighted and fused with the real physical road network graph, thereby balancing prior physical knowledge and data-implicit dependencies.

    \item \textbf{Dynamic memory bank mechanism}: A learnable memory bank is constructed to store typical traffic pattern prototypes, and an attention mechanism is used to generate personalized pattern representations for each sensor; simultaneously, a dynamic graph updater is designed to adaptively adjust the graph structure during decoding to cope with real-time changes.
\end{itemize}

\section{Related Work}
\label{sec:related_work}

\subsection{Spatio-temporal Traffic Prediction}
Early traffic prediction research primarily relied on statistical models, such as Historical Average (HA), Autoregressive Integrated Moving Average (ARIMA) models, and their vector variants (VAR) \cite{zivot2006vector}. These methods are typically based on linear assumptions and struggle to fit the complex nonlinear characteristics of traffic flow. With the advancement of machine learning techniques, methods like Support Vector Regression (SVR) and Random Forests were introduced, but they still have limitations when dealing with high-dimensional spatio-temporal dependencies.

The rise of deep learning has brought revolutionary progress to this field. CNN-based methods \cite{zhang2017deep} often process urban areas by dividing them into grid images, but this Euclidean space approximation ignores the actual road network topology. RNN-based methods (e.g., LSTM, GRU) \cite{cho2014learning} are good at handling sequence data but are less efficient at capturing long-range temporal dependencies. Recently, the Transformer architecture based on self-attention mechanisms has shown excellent performance in long-sequence modeling \cite{vaswani2017attention}, but it also faces challenges of high computational complexity ($O(L^2)$) and strong reliance on large-scale data.

\subsection{Graph Neural Networks in Traffic Prediction}
The proposal of Graph Convolutional Networks (GCNs) \cite{kipf2016semi} provides a powerful tool for processing non-Euclidean spatial data. In the traffic field, DCRNN \cite{li2018diffusion} pioneered the combination of diffusion convolution processes on graphs with RNNs; STGCN \cite{yu2018spatio} used Chebyshev polynomials to approximate spectral graph convolution for efficient computation. Subsequent Graph WaveNet \cite{wu2019graph} introduced an adaptive adjacency matrix to supplement missing connections, and ASTGCN \cite{guo2019attention} further integrated spatio-temporal attention mechanisms.

However, the above methods still have shortcomings: (1) Excessive reliance on a single predefined graph structure cannot reflect the dynamic nature of traffic flow over time; (2) Graph convolution layers share parameters across all nodes, ignoring the spatio-temporal heterogeneity among traffic nodes.

\subsection{Adaptive Graph Generation and Memory Networks}
To break free from dependency on prior graph structures, AGCRN \cite{bai2020adaptive} proposed a Node Adaptive Parameter Learning (NAPL) strategy, achieving fully data-driven graph generation. MegaCRN \cite{jiang2022megacrn} introduced Memory Networks to store and retrieve historical traffic patterns. The GEnSHIN proposed in this paper combines the advantages of the above methods. Through an asymmetric embedding graph generation mechanism and a dynamic memory bank, it effectively learns latent spatio-temporal topological structures while preserving the physical constraints of the real road network.

\section{Methodology}
\label{sec:methodology}

\subsection{Problem Definition}
Traffic flow prediction aims to forecast future traffic states based on historical observation data. Let there be $N$ traffic sensors (nodes), with observations at each time step $t$ being $\mathbf{X}_t \in \mathbb{R}^{N \times C}$, where $C$ is the feature dimension (e.g., flow, speed). Given data from the past $T$ time steps $\mathbf{X} = [\mathbf{X}_{t-T+1}, \dots, \mathbf{X}_t] \in \mathbb{R}^{T \times N \times C}$, the goal is to predict values for the next $\tau$ time steps $\mathbf{Y} = [\mathbf{X}_{t+1}, \dots, \mathbf{X}_{t+\tau}] \in \mathbb{R}^{\tau \times N \times C}$.

Model the traffic network as a graph $\mathcal{G} = (\mathcal{V}, \mathcal{E})$, where $\mathcal{V}$ is the node set ($|\mathcal{V}| = N$), $\mathcal{E}$ is the edge set, and $\mathbf{A} \in \{0,1\}^{N \times N}$ is the adjacency matrix. The prediction problem can be formally defined as:
\begin{equation}
    \mathbf{Y} = \mathcal{F}_{\theta}(\mathbf{X}, \mathbf{A})
\end{equation}
where $\mathcal{F}_{\theta}$ is the proposed GEnSHIN model, and $\theta$ are the model parameters.

\subsection{Model Architecture}
\begin{figure*}[htbp]
\centering
\includegraphics[scale=0.19]{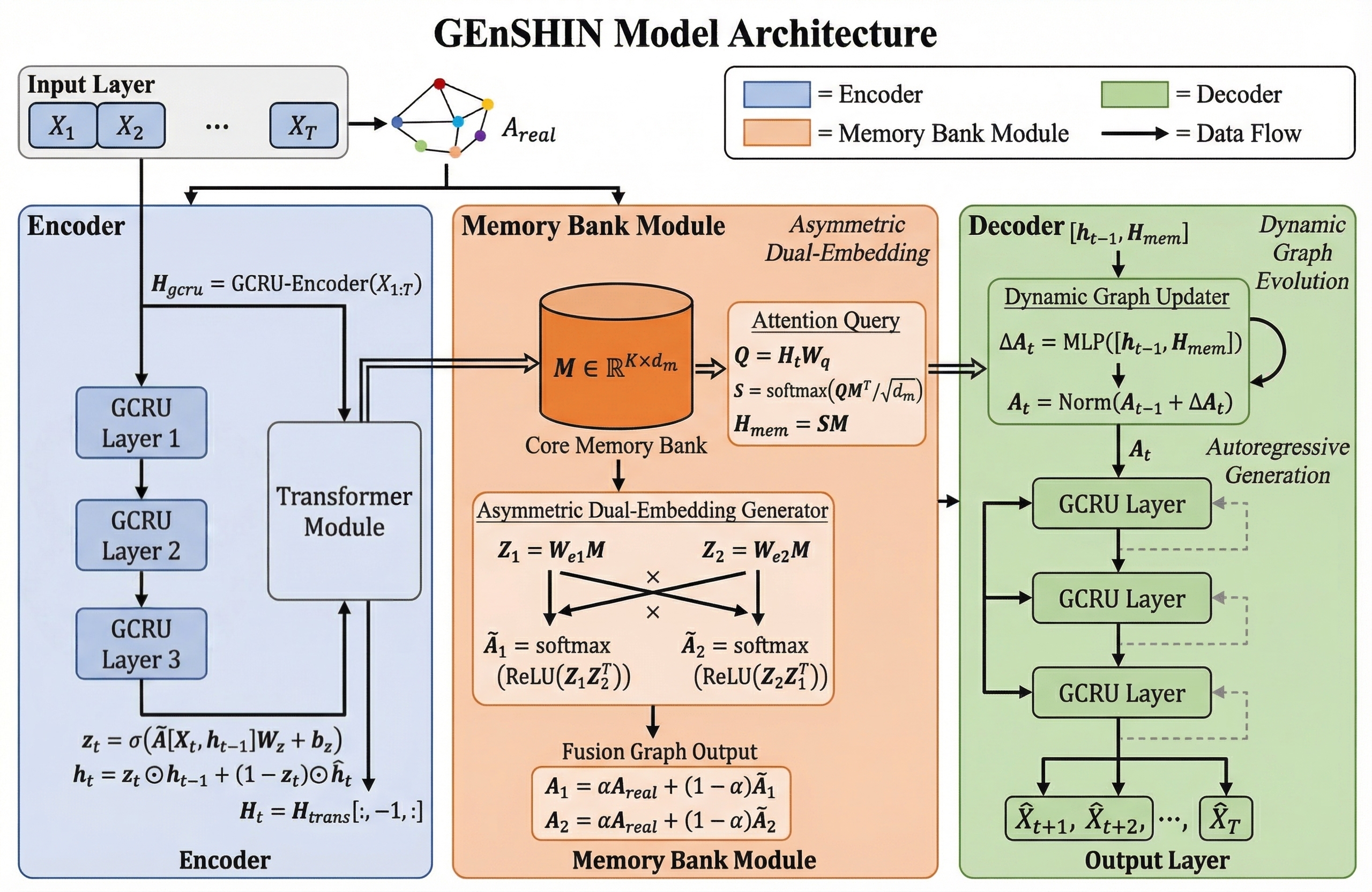}
\caption{GEnSHIN Model Architecture Diagram}
\label{fig:model_architecture}
\end{figure*}

As shown in Figure \ref{fig:model_architecture}, GEnSHIN consists of three parts: an encoder, a memory bank module, and a decoder. The encoder extracts spatio-temporal features through GCRU units and Transformers; the memory bank module stores and retrieves traffic pattern prototypes; the decoder performs multi-step prediction based on enhanced representations and a lightweight graph updater.

\subsubsection{Attention-Enhanced GCRU}
In the DGCRN \cite{li2022dynamic} and AGCRN \cite{bai2020adaptive} networks, the idea of replacing the weight matrices in the linear transformations of RNNs with graph convolutions cleverly integrates spatial information represented by the graph with time series. In our experiment, we reference this idea and set up a similar GCRU unit:
\begin{align}
    \vect{z}_t &= \sigma\left(\tilde{\vect{A}} [\vect{X}_t, \vect{h}_{t-1}] \vect{W}_z + \vect{b}_z\right) \tag{1} \\
    \vect{r}_t &= \sigma\left(\tilde{\vect{A}} [\vect{X}_t, \vect{h}_{t-1}] \vect{W}_r + \vect{b}_r\right) \tag{2} \\
    \hat{\vect{h}}_t &= \tanh\left(\tilde{\vect{A}} [\vect{X}_t, \vect{r}_t \odot \vect{h}_{t-1}] \vect{W}_h + \vect{b}_h\right) \tag{3} \\
    \vect{h}_t &= \vect{z}_t \odot \vect{h}_{t-1} + (1 - \vect{z}_t) \odot \hat{\vect{h}}_t \tag{4}
\end{align}
  where $\tilde{\vect{A}}$ is the normalized adjacency matrix, and $[\cdot,\cdot]$ denotes the concatenation operation.
However, we believe that while the GCRU unit captures local spatio-temporal dependencies very well, its ability to maintain long-term dependencies is not perfect due to the limitations of the RNN architecture. Therefore, we reference the approach in Traffic STGNN \cite{wang2020traffic} and set up a Transformer module to enhance the capture of global temporal dependencies.

During encoding, we use the hidden states output by GCRU as input and establish global temporal dependencies through the Transformer module. The Transformer output at the last time step is used as the final encoding result of the encoder.
\begin{align}
    \vect{H}_{\text{gcru}} &= \text{GCRU-Encoder}(\vect{X}_{1:T}) \tag{5} \\
    \vect{H}_{\text{trans}} &= \text{Transformer}(\vect{H}_{\text{gcru}}) \tag{6} \\
    \vect{H}_t &= \vect{H}_{\text{trans}}[:, -1, :] \tag{7}
\end{align}

\subsubsection{Memory-Enhanced Asymmetric Graph Structure Learning}
Adaptive graph structure learning has been validated by many teams as effective in capturing latent spatio-temporal dependencies in spatio-temporal graph data \cite{bai2020adaptive,li2022dynamic}. However, we believe this approach of completely discarding real graph information is too aggressive; furthermore, in these studies, the learned graph structure is generated using a single embedding vector $\vect{E} \in \reals^{N \times d}$ via $\text{softmax}(\text{ReLU}(\vect{E}\vect{E}^\top))$, resulting in a symmetric graph. But in real traffic networks, graphs are often asymmetric. For example, during morning peak hours, the busyness from suburbs to the city center and from the city center to suburbs is necessarily different. To simultaneously leverage the prior knowledge of the real road network and capture latent asymmetric spatio-temporal dependencies in the data, we design a memory bank-driven graph structure learning module. It consists of a traffic pattern memory bank $\mathcal{M} \in \reals^{K \times d_m}$ (which stores $K$ prototypes of dimension $d_m$) whose effectiveness was validated in MegaCRN \cite{jiang2022megacrn}, and two learnable node-pattern association matrices $\vect{W}_{e1}, \vect{W}_{e2} \in \reals^{N \times K}$. The model learns two embedding representation vectors:
\begin{align}
    \vect{Z}_1 &= \vect{W}_{e1} \mathcal{M} \quad \in \reals^{N \times d_m} \tag{8} \\
    \vect{Z}_2 &= \vect{W}_{e2} \mathcal{M} \quad \in \reals^{N \times d_m} \tag{9}
\end{align}
Subsequently, two asymmetric graph adjacency matrices based on historical patterns are generated via alternating inner product and Softmax normalization:
\begin{align}
    \tilde{\vect{A}}_1 &= \text{softmax}(\text{ReLU}(\vect{Z}_1 \vect{Z}_2^\top)) \tag{10} \\
    \tilde{\vect{A}}_2 &= \text{softmax}(\text{ReLU}(\vect{Z}_2 \vect{Z}_1^\top)) \tag{11}
\end{align}
Finally, they are weighted and fused with the real road network matrix $\vect{A}_{\text{real}}$ to obtain the static fused graph used by the encoder:
\begin{align}
    \vect{A}_1 &= \alpha \vect{A}_{\text{real}} + (1 - \alpha) \tilde{\vect{A}}_1 \tag{12} \\
    \vect{A}_2 &= \alpha \vect{A}_{\text{real}} + (1 - \alpha) \tilde{\vect{A}}_2 \tag{13}
\end{align}
where $\alpha$ is a learnable fusion weight. Simultaneously, the encoder's output state $\vect{H}_t$ queries the memory bank via an attention mechanism to obtain node-specific pattern-enhanced representations:
\begin{align}
    &\vect{Q} = \vect{H}_t \vect{W}_q, \quad \vect{S} = \text{softmax}(\frac{\vect{Q} \mathcal{M}^\top}{\sqrt{d_m}}) \tag{14} \\
    &\vect{H}_{\text{mem}} = \vect{S} \mathcal{M} \tag{15}
\end{align}
This enhanced representation $\vect{H}_{\text{mem}}$ is concatenated with the encoder output $\vect{H}_t$ as the initial input to the decoder. This pattern enhancement was considered an effective method by the authors of MegaCRN \cite{jiang2022megacrn}. Furthermore, $\vect{H}_{\text{mem}}$ will provide the basis for dynamic graph updating.

\subsubsection{Dynamic Graph Updater}
In the decoding phase, we reference the idea from MegaCRN \cite{jiang2022megacrn} for auto-regressive decoding, but with slight differences in graph generation and updating. We believe the future traffic network structure should not be static but should dynamically change over time. Therefore, unlike MegaCRN \cite{jiang2022megacrn} which multiplies the learned embedding vector with the pattern memory to obtain a static, unchanging graph structure, we use the decoder's current hidden state $\vect{h}_{t-1}$ concatenated with the memory-enhanced representation $\vect{H}_{\text{mem}}$ to dynamically adjust the graph structure via a lightweight network, better adapting to predicting the current traffic state:
\begin{align}
    \Delta \vect{A}_t &= \text{MLP}([\vect{h}_{t-1}, \vect{H}_{\text{mem}}]) \tag{16} \\
    \vect{A}_t &= \text{Norm}(\vect{A}_{t-1} + \Delta \vect{A}_t) \tag{17}
\end{align}
where MLP is a Multi-Layer Perceptron, and Norm denotes row normalization. $\vect{A}_{t-1}$ is initially one of the static fused graphs obtained by the encoder.

\subsection{Training Objectives}
The model's total loss function consists of three parts: the prediction task loss, pattern consistency loss, and contrastive loss:
\begin{equation}
    \mathcal{L} = \mathcal{L}_{\text{task}} + \lambda_1 \mathcal{L}_{\text{consistency}} + \lambda_2 \mathcal{L}_{\text{contrast}} \tag{18}
\end{equation}
where $\lambda_1$ and $\lambda_2$ are hyperparameters balancing the weights of each loss term.

\noindent\textbf{Prediction Task Loss}: Mean Absolute Error (MAE) is used to measure the deviation between predicted and true values:
\begin{equation}
    \mathcal{L}_{\text{task}} = \frac{1}{\tau N} \sum_{i=1}^{\tau} \sum_{j=1}^{N} |\hat{\vect{X}}_{t+i,j} - \vect{X}_{t+i,j}| \tag{19}
\end{equation}

\noindent\textbf{Pattern Consistency Loss}: To ensure the encoded state remains consistent with its primarily activated memory pattern, we minimize the distance between the query vector $\vect{Q}$ and the most relevant memory prototype $\vect{M}^+$:
\begin{equation}
    \mathcal{L}_{\text{consistency}} = \|\vect{Q} - \vect{M}^+\|^2 \tag{20}
\end{equation}

\noindent\textbf{Contrastive Loss}: A variant of Triplet Loss is used to bring the query vector closer to the most relevant prototype $\vect{M}^+$ while pushing it away from the second most relevant prototype $\vect{M}^-$:
\begin{equation}
    \mathcal{L}_{\text{contrast}} = \max\left(0, \|\vect{Q} - \vect{M}^+\|^2 - \|\vect{Q} - \vect{M}^-\|^2 + \gamma \right) \tag{21}
\end{equation}
where $\gamma > 0$ is a margin parameter.

\section{Experiments}
\label{sec:experiments}

\subsection{Datasets and Experimental Settings}
We conduct experiments on a public traffic dataset \cite{li2022dynamic}: METR-LA (Los Angeles freeways, 207 sensors). Data is aggregated at 5-minute intervals, using the past 60 minutes of data to predict the next 60 minutes (12 time steps). The dataset is first split in chronological order, then divided into training, validation, and test sets in a 7:1:2 ratio.

Evaluation metrics include: Mean Absolute Error (MAE), Root Mean Square Error (RMSE), and Mean Absolute Percentage Error (MAPE). All experiments are conducted on an NVIDIA A800-sxm80g using PyTorch. We use the AdamW optimizer with an initial learning rate of 0.001, weight decay of 1e-4, and a batch size of 64. The model is trained for a total of 100 epochs, with an early stopping strategy (patience=20) to prevent overfitting. The maximum gradient norm for clipping is set to 5.0 to ensure training stability.

The specific configuration of GEnSHIN is as follows: The encoder and decoder each contain 5 layers of GCRU units with a hidden dimension of 128; the Chebyshev polynomial order is 3; the memory bank contains 20 memory prototypes, each with a dimension of 64; the Transformer module consists of 2 encoder layers with 4 attention heads; the intermediate layer dimension of the dynamic graph updater is 128. In the loss function, the task loss weight is 1.0, the consistency loss weight is 0.01, the contrastive loss weight is 0.01, and the margin parameter for contrastive loss is set to 1.0.

\subsection{Comparison Methods}
To evaluate model performance, we select representative baseline methods from multiple categories for comparison, as follows:

\noindent \textbf{Traditional Methods}
\begin{itemize}
    \item \textbf{Historical Average (HA)} \cite{li2022dynamic}: A classic statistical baseline method that uses the average of historical data from the same period as the prediction. This method is computationally simple but cannot capture complex spatio-temporal dynamic dependencies in traffic data.
\end{itemize}

\noindent \textbf{Deep Learning Methods}
\begin{itemize}
    \item \textbf{Spatio-Temporal Transformer Network (STTN)} \cite{xu2020spatial}: A deep learning model based on the Transformer architecture that models global dependencies in both temporal and spatial dimensions through self-attention mechanisms.
\end{itemize}

\noindent \textbf{Graph Neural Network Methods}
\begin{itemize}
    \item \textbf{Spatio-Temporal Graph Convolutional Networks (STGCN)} \cite{yu2018stgcn}: Combines Graph Convolutional Networks (GCN) with temporal convolutions, using spectral graph convolution to capture spatial dependencies and 1D causal convolutions to extract temporal features.
    \item \textbf{Diffusion Convolutional Recurrent Neural Network (DCRNN)} \cite{li2022dynamic}: Models the traffic road network as a diffusion process, embedding diffusion graph convolutions within the Recurrent Neural Network (RNN) framework to simultaneously learn spatio-temporal dynamics.
    \item \textbf{Adaptive Graph Convolutional Recurrent Network (AGCRN)} \cite{bai2020adaptive}: Dynamically learns latent graph representations from data through node adaptive parameter learning and adaptive graph structure learning.
    \item \textbf{Coupled Hierarchical Convolutional Recurrent Neural Network (CCRNN)} \cite{ye2021coupled}: Uses adaptive coupled hierarchical graph convolution and multi-level aggregation modules to improve temporal convolutional modules, extracting spatio-temporal dependency information from multiple perspectives.
\end{itemize}

\noindent The above methods model the spatio-temporal correlations in traffic prediction from different angles, providing a multi-level performance benchmark for this study.

\subsection{Main Results}
Table \ref{tab:main_results} shows the 12-step prediction results on the METR-LA dataset. GEnSHIN achieves the best performance on MAE and MAPE and is highly competitive on RMSE. Overall, our model's performance is outstanding.

\begin{table}[htbp]
\centering
\caption{Forecasting performance on METR-LA}
\label{tab:main_results}
\begin{tabular}{lccc}
\toprule
Model & MAE & RMSE & MAPE \\
\midrule
HA\cite{li2022dynamic} & 4.16 & 7.80 & 13.00\% \\
STGCN\cite{yu2018stgcn} & 4.59 & 9.40 & 12.70\% \\
DCRNN\cite{li2022dynamic} & 3.60 & 7.59 & 10.50\% \\
STTN\cite{xu2020spatial} & 3.60 & 7.60 & 10.16\% \\
AGCRN\cite{bai2020adaptive} & 3.68 & \textbf{7.56} & 10.46\% \\
CCRNN\cite{ye2021coupled} & 3.73 & 7.65 & 10.59\% \\
\midrule
\textbf{GEnSHIN (Ours)} & \textbf{3.60} & 7.69 & \textbf{9.06\%}\\
\bottomrule
\end{tabular}
\end{table}

\subsection{Ablation Study}
To verify the effectiveness of each module, we design the following variants:
\begin{itemize}
    \item \textbf{w/o Transformer}: Remove the Transformer module, using the hidden states output by GCRU as the encoding.
    \item \textbf{w/o Dual Embed}: Use a single embedding vector to generate the graph, verifying the effectiveness of asymmetric dual embedding.
    \item \textbf{w/o Memory}: Remove the memory bank module, not using pattern enhancement.
    \item \textbf{w/o Dynamic Graph}: Use a static graph during the decoding phase, not updating the graph structure.
    \item \textbf{w/o Real Graph}: Do not use the real road network graph, only use the learned graph.
\end{itemize}

The results are shown in Table \ref{tab:ablation}. The full model performs best, with each module contributing positively, demonstrating the effectiveness of our design. Among them, ablating the Transformer and dynamic graph modules has the greatest impact on model performance, highlighting the importance of capturing global dependencies and dynamic adaptation in spatio-temporal prediction.
\begin{table}[htbp]
\centering
\caption{Ablation Study Results (METR-LA Dataset)}
\label{tab:ablation}
\begin{tabular}{lccc}
\toprule
Model & MAE & RMSE & MAPE(\%) \\
\midrule
\textbf{Whole Model} & \textbf{3.60} & \textbf{7.69} & \textbf{9.06}\\
w/o Transformer & 3.88 & 9.50 & 9.85 \\
w/o Dual Embed & 3.74 & 9.46 & 9.71 \\
w/o Memory & 3.67 & 9.32 & 9.43 \\
w/o Dynamic Graph & 3.76 & 9.39 & 9.54 \\
w/o Real Graph & 3.64 & 9.24 & 9.08 \\

\bottomrule
\end{tabular}
\end{table}

\subsection{Visualization Analysis}
\begin{figure}[htbp]
\centering
\includegraphics[scale=0.28]{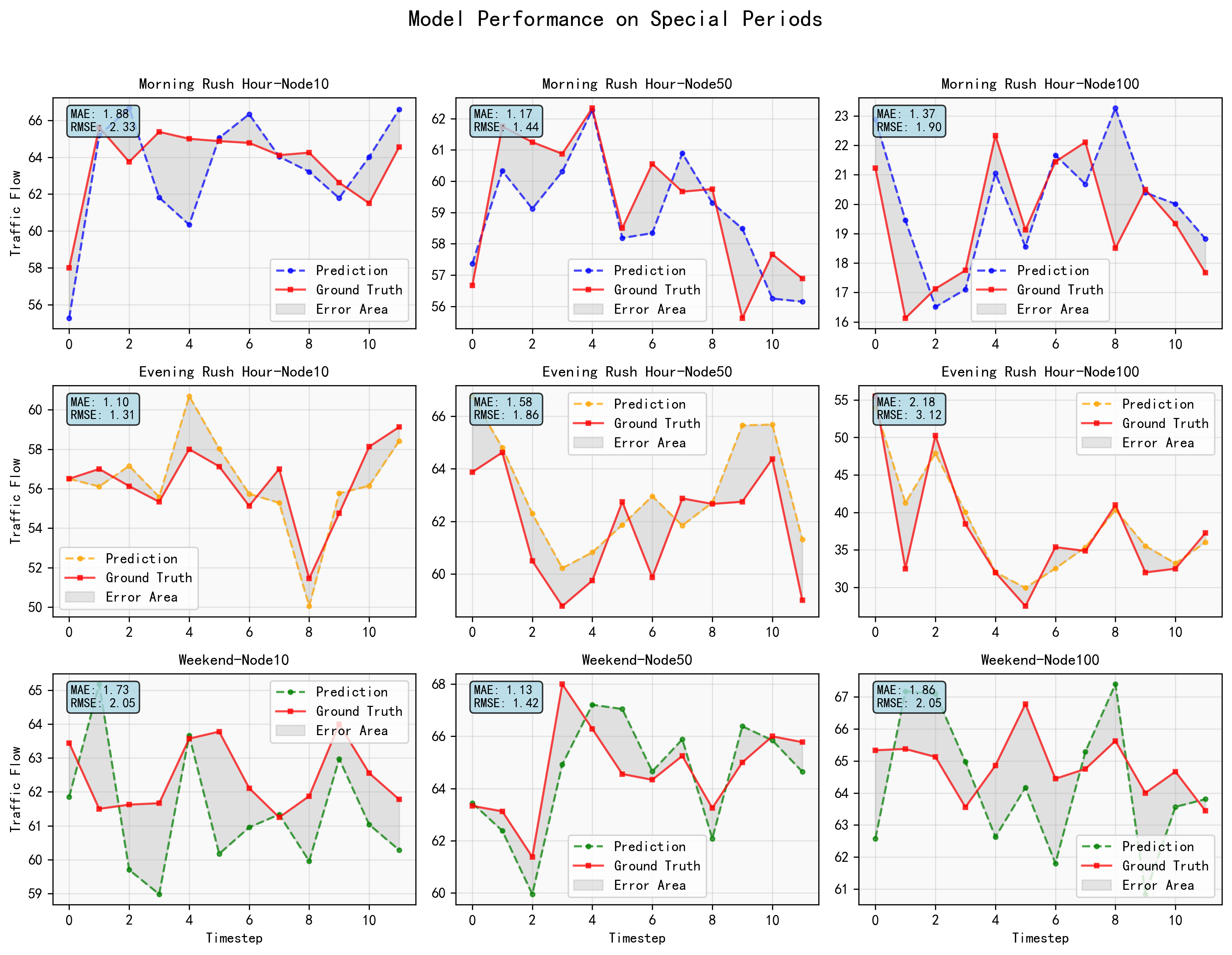}
\caption{Traffic Flow Prediction Results vs. True Values During Special Time Periods}
\label{fig:pred_effect}
\end{figure}

Figure \ref{fig:pred_effect} shows the comparison curves between the model's prediction results and the true values for three sensor nodes during three special time periods selected based on timestamps from the METR-LA dataset: morning peak, evening peak, and weekend. It can be seen that in these special, busier traffic periods, GEnSHIN still demonstrates relatively good fitting ability. The predicted curves generally match the true curves in terms of trend and magnitude of change.

\begin{figure}[htbp]
\centering
\includegraphics[scale=0.58]{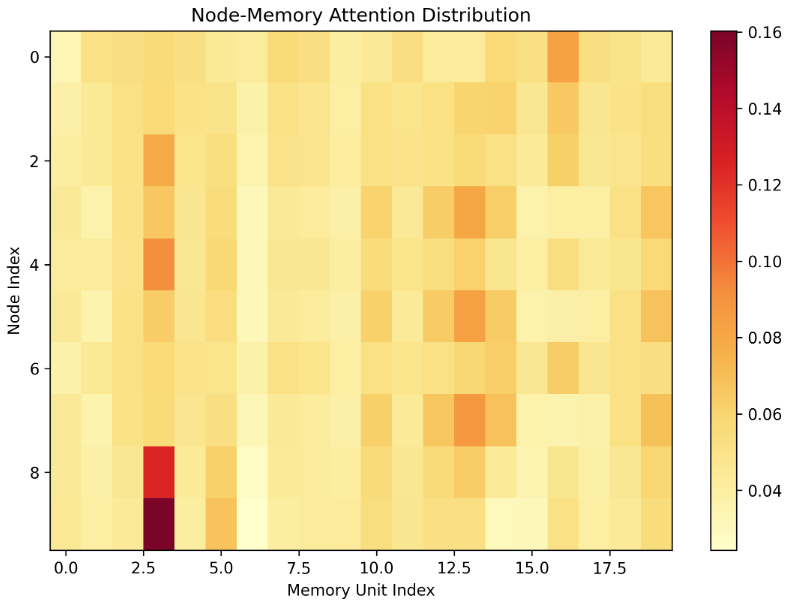}
\caption{Traffic Pattern Attention Heatmap (First 10 Nodes)}
\label{fig:att}
\end{figure}

Figure \ref{fig:att} is a heatmap of the attention scores for the first ten nodes in the first batch of the test set towards the traffic pattern memory bank prototypes. The horizontal axis is the traffic pattern memory prototype index (it should not actually be displayed as floating-point numbers, but after several attempts it still shows as such), the vertical axis is the node index, and each coordinate point represents the attention level of the current node towards the current pattern during querying. It can be observed that the selected nodes show grouped attention patterns towards different prototypes, roughly divided into three situations, which seems to reflect that nodes in the current batch share common patterns of focus. Also, observing the heatmap of attention levels for each memory prototype unit reveals that there are a few prototypes that are highly attended and many prototypes that receive little attention. We believe this validates the effectiveness of the pattern memory bank—the current batch successfully found the patterns most similar to their own in the pattern memory bank, rather than attending to all patterns uniformly.

\section{Conclusion}
\label{sec:conclusion}

This paper proposed a novel Graphical Enhanced Spatiotemporal Hierarchical Inference Network (GEnSHIN) for traffic flow prediction. Through the attention-enhanced GCRU unit, asymmetric dual-embedding graph learning, a lightweight dynamic graph updater during decoding, and a pattern memory bank module, our model can effectively capture the complex spatio-temporal dependencies in traffic data. Experimental results on the public METR-LA dataset show that GEnSHIN has excellent performance. We validated the effectiveness of each module through ablation studies, demonstrating the rationality of our initial design. Furthermore, we presented visualization results showing the model's outstanding prediction performance during special, busy traffic periods and the working mode of the pattern memory bank during actual inference.

Future work will focus on the following directions: (1) Validating the model's performance on more spatio-temporal traffic datasets; (2) Extending the model to handle spatio-temporal graph data in other domains; (3) Exploring more efficient module implementations to reduce computational complexity.

\section*{Acknowledgments}
We thank the Supercomputing Center of Beijing Normal University Zhuhai Campus for providing computational resources to support this experiment. We thank all the author teams of the referenced papers for providing valuable references.

\section*{Code Availability}
The code for this paper is open-sourced at: \\
\url{https://github.com/airyuanshen/GEnSHIN}


\appendix
\section{Appendix}
\subsection{Pseudocode}
\begin{algorithm}[htbp]
\caption{GEnSHIN Training Process}
\begin{algorithmic}[1]
\REQUIRE Training data $\mathcal{D}$, graph structure $\vect{A}_{\text{real}}$, hyperparameters
\ENSURE Trained model parameters $\theta$
\STATE Initialize model parameters $\theta$, memory bank $\mathcal{M}$
\FOR{epoch = 1 to $E$}
    \FOR{batch $(\vect{X}, \vect{Y})$ in $\mathcal{D}$}
        \STATE // Forward propagation
        \STATE Encode: $\vect{H}_t \leftarrow \text{Encoder}(\vect{X}, \vect{A}_{\text{real}})$
        \STATE Memory query: $\vect{H}_{\text{mem}} \leftarrow \text{MemoryBank}(\vect{H}_t)$
        \STATE Predict: $\hat{\vect{Y}} \leftarrow \text{Decoder}(\vect{H}_t, \vect{H}_{\text{mem}})$

        \STATE // Loss calculation
        \STATE $\mathcal{L}_{\text{pred}} \leftarrow \text{MAE}(\hat{\vect{Y}}, \vect{Y})$

        \STATE$\mathcal{L}_{\text{consistency}} \leftarrow\|\vect{Q} - \vect{M}^+\|^2 $\STATE$\mathcal{L}_{\text{contrast}} \leftarrow \max\left(0, \|\vect{Q} - \vect{M}^+\|^2 - \|\vect{Q} - \vect{M}^-\|^2 + \gamma \right) $

        \STATE $\mathcal{L} \leftarrow \mathcal{L}_{\text{task}} + \lambda_1 \mathcal{L}_{\text{consistency}} + \lambda_2 \mathcal{L}_{\text{contrast}} $
        \STATE // Backpropagation
        \STATE $\theta \leftarrow \theta - \eta \nabla_{\theta} \mathcal{L}$
    \ENDFOR
\ENDFOR
\end{algorithmic}
\end{algorithm}

\end{document}